%% file: arxiv.tex
\documentclass{article}

\usepackage{arxiv}

\usepackage[utf8]{inputenc} 
\usepackage[T1]{fontenc}    
\usepackage{hyperref}       
\usepackage{url}            
\usepackage{booktabs}       
\usepackage{amsfonts}       
\usepackage{nicefrac}       
\usepackage{microtype}      
\usepackage{graphicx}
\usepackage{natbib}
\usepackage{doi}

\usepackage{adjustbox}
\usepackage{xcolor}         
\usepackage{amsmath}
\usepackage{amssymb}
\usepackage{xspace}
\usepackage{color}

\usepackage{marvosym}
\usepackage{authblk}

\makeatletter
\renewcommand\@author{%
  \ifx\AB@affillist\AB@empty
    \AB@author
  \else
    \ifnum\value{affil}>\value{Maxaffil}
      \def\rlap##1{##1}%
      \AB@authlist\\[\affilsep]%
      \parbox{\textwidth}{\raggedright\AB@affillist}%
    \else
      \AB@authors
    \fi
  \fi}
\makeatother

\input{preamble}

\newif\ifAIAI
\AIAIfalse

\usepackage{authblk}

\title{\cycleGANli: Object-Centric Representation Learning using Cycle Consistent Generative Adversarial Networks}

\date{}

\author[a,$\dag$]{Jo\"el K\"uchler}
\author[b]{Ellen van Maren}
\author[a]{Vaiva Vasiliauskait\.e}
\author[a]{Katarina Vuli\'c}
\author[c]{Reza Abbasi-Asl}
\author[a,d,e,$\dag$,\Letter]{Stephan J. Ihle}

\affil[a]{Laboratory of Biosensors and Bioelectronics, Institute for Biomedical Engineering, University and ETH Zurich, Gloriastrasse 37/39, Zurich, 8092, Switzerland}                      
\affil[b]{Department of Neurology, Insel Gruppe, Bern, Switzerland}                           
\affil[c]{Department of Neurology, Department of Bioengineering and Therapeutic Sciences, University of California, San Francisco, USA}                             
\affil[d]{Department of Neurobiology, University of Chicago, 951 E 58th St, Chicago, 60637, IL, USA}                                      
\affil[e]{Department of Physics, University of Chicago, 929 E 57th St, Chicago, 60637, IL, USA}
\affil[$\dag$]{These authors contributed equally to this work.}
\affil[\Letter]{Corresponding author: \texttt{ihles@uchicago.edu}}
\date{}

\renewcommand{\headeright}{}
\renewcommand{\undertitle}{}
\renewcommand{\shorttitle}{}

\hypersetup{
    pdftitle={ORGAN: Object-centric Representation Learning using Cycle Consistent Generative Adversarial Networks},
    pdfauthor={Jo\"el K\"uchler, Ellen van Maren, Vaiva Vasiliauskait\.e, Katarina Vuli\'c, Reza Abbasi-Asl, Stephan J. Ihle},
    pdfkeywords={CycleGAN, Generative models, Object-centric representation learning, Representation learning.},
}

\usepackage{xr-hyper} 

\makeatletter
\newcommand*{\addFileDependency}[1]{
  \typeout{(#1)}
  \@addtofilelist{#1}
  \IfFileExists{#1}{}{\typeout{No file #1.}}
}

\makeatletter
\newcommand{\@noticestring}{%
    \emph{Proceedings of the 22nd International Conference on Artificial Intelligence Applications and Innovations, Chania, Greece, 2026. IFIP AICT volume 792. Copyright 2027 by the author(s).}
}
\makeatother

\newcommand*{\myexternaldocument}[1]{%
    \externaldocument{#1}%
    \addFileDependency{#1.tex}%
    \addFileDependency{#1.aux}%
}
\makeatother
\myexternaldocument{Arxiv/arxiv_SI}

\begin{document}
\maketitle
\makeatletter\@notice\makeatother
\input{sec/0_abstract}   

\input{sec/1_intro}
\input{sec/2_related_work}

\input{sec/3_methodology}
\input{sec/4_experiments}

\input{sec/5_conclusion}
\input{sec/6_acknowledgements}
\bibliographystyle{unsrtnat}
\bibliography{main}


\end{document}

%% file: preamble.tex
\newcommand{\ie}{\textit{i.e.}}

\newcommand{\cycleGANli}{ORGAN\xspace}
\newcommand{\sprites}{\textit{Sprites}\xspace}
\newcommand{\mnist}{\textit{Multi-MNIST}\xspace}

\newcommand{\tetris}{\textit{Tetrominoes}\xspace}
\newcommand{\cells}{\textit{Cells}\xspace}

\newcommand{\fig}[2]{Fig.~\ref{#1}#2}

%% file: sec/0_abstract.tex
\begin{abstract}
    Although data generation is often straightforward, extracting information from data is more difficult. 
    Object-centric representation learning can extract information from images in an unsupervised manner. It does so by segmenting an image into its subcomponents: the objects. Each object is then represented in a low-dimensional latent space that can be used for downstream processing. 
    Object-centric representation learning is dominated by autoencoder architectures (AEs).
    Here, we present \cycleGANli, a novel approach for object-centric representation learning, which is based on cycle-consistent Generative Adversarial Networks instead. 
    We show that it performs similarly to other state-of-the-art approaches on synthetic datasets, while at the same time being the only approach tested here capable of handling more challenging real-world datasets with many objects and low visual contrast. Complementing these results, \cycleGANli creates expressive latent space representations that allow for object manipulation.
    Finally, we show that \cycleGANli scales well both with respect to the number of objects and the size of the images, giving it a unique edge over current state-of-the-art approaches.
    \keywords{CycleGAN \and Generative models \and Object-centric representation learning \and Representation learning.}
\end{abstract}

%% file: sec/1_intro.tex
\section{Introduction}
\label{sec:intro}

Image datasets are ever-increasing in size. Yet, labeling their content remains costly. As a consequence, unsupervised object detection in images is a pressing need in the field of machine learning \cite{pervez2022differentiable,locatello2020object}. Object-centric representation learning is a powerful tool towards this goal. It focuses on estimating the presence, location, and properties of objects of a given scene in an unsupervised fashion. As such, it is capable of separating a scene into multiple objects, which are condensed into a latent representation.

\begin{figure}[t]
    \centering
    \ifAIAI\else \includegraphics[width=0.9\linewidth]{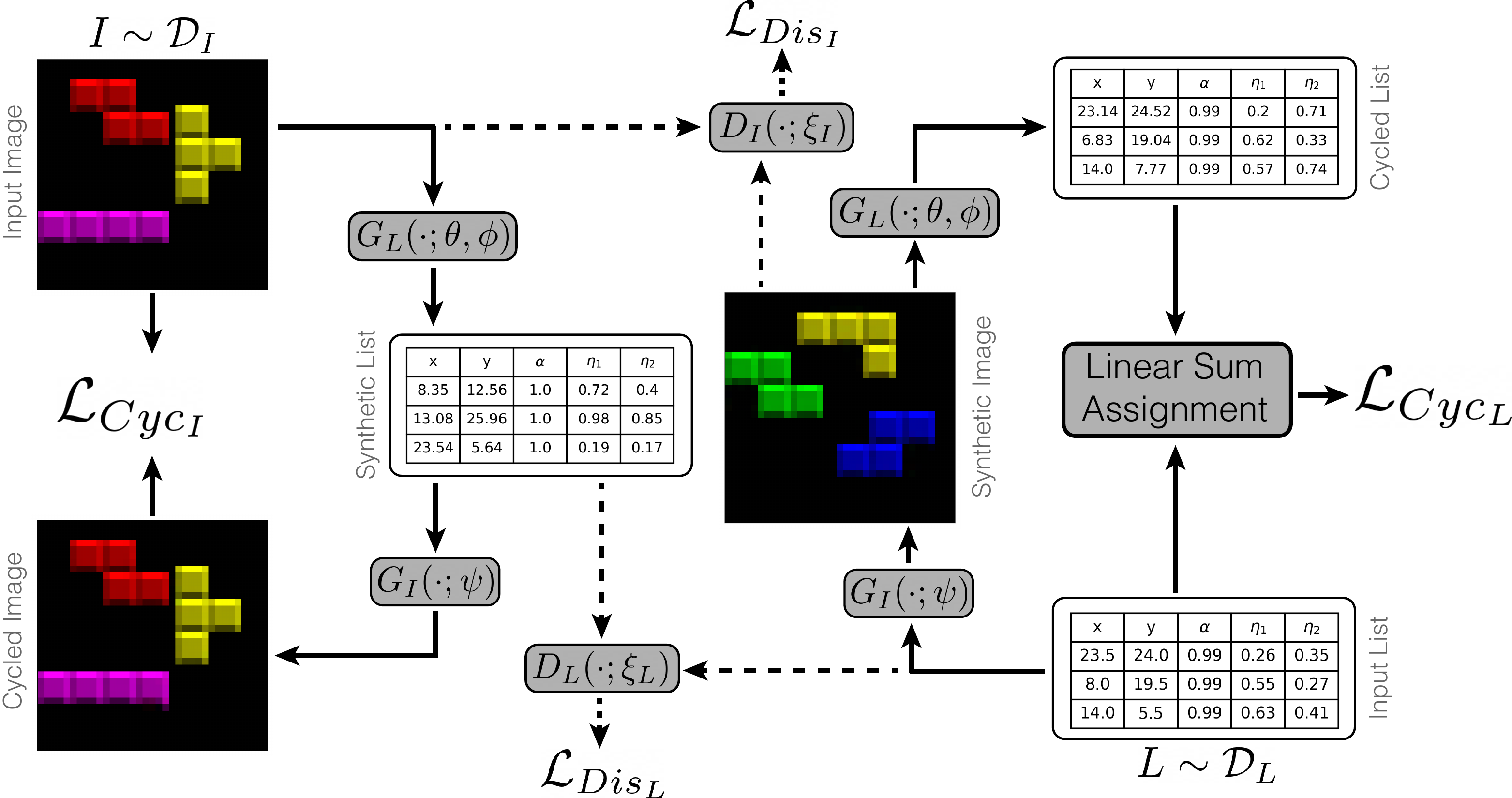} \fi
    \caption{Model architecture and training losses. \cycleGANli is based on the Cycle-Consistent Generative Adversarial Network (CycleGAN) architecture \cite{zhu2017unpaired}, adapted to use an image (left) and a list (right) domain. Two GANs are used to transform between the two domains. By cycling through both generators, the cycle-consistency losses $\mathcal{L}_{Cyc_L}$ and $\mathcal{L}_{Cyc_I}$ can be enforced.}
    \label{fig:0_train}
\end{figure}


Object-centric representations are similar to the way humans reason and make predictions about the physical world \cite{battaglia2013simulation,lake2017building,kubricht2017intuitive,spelke1992origins}. Therefore, it is no wonder that these types of representations are deemed more interpretable \cite{mascharka2018transparency}. Object-centric representations can be recovered using classical approaches, such as wavelet representations \cite{papageorgiou1998general} and template matching \cite{betke1995fast}, as well as using deep learning. Unsupervised techniques are desirable, as they do not require labels but instead focus on intrinsic biases in the data. Two well-known unsupervised classes of architectures that can be used to create representations are variational auto-encoders (VAEs) \cite{kingma2013auto} and generative adversarial networks (GANs) \cite{goodfellow2014generative}. However, there are also other unsupervised approaches with similar capabilities \cite{kipf2019contrastive}.

Usually, object-centric representation learning is done with VAEs. Popular representation learning approaches have been benchmarked by Yang and Yang \cite{yang2024benchmarking}. In general, these approaches can be divided into spatial-attention models and scene-mixture models \cite{lin2020space}. For spatial-attention models, the location and object properties are explicitly modeled. Scene-mixture models, on the other hand, decompose a scene into multiple layers, where each pixel is assigned to one layer \cite{greff2017neural,engelcke2021genesis}. The layers are also called slots \cite{locatello2020object}. Over time, object-centric representation learning has been advanced to the point that it is now possible to learn representations of complex real-world data \cite{seitzer2022bridging}.

GANs are another powerful tool for image generation. Given a dataset, a generator network learns to create new data samples such that they match the training data distribution. A discriminator network tries to distinguish between real and fake data and teaches the generator. Their focus lies in generating images as a composition of multiple objects \cite{greff2020binding}. 

In addition to being used for directly inferring object representations, GANs have also been used to indirectly infer object properties. Of particular interest to us are Cycle-Consistent Generative Adversarial Networks (CycleGANs) \cite{zhu2017unpaired}, which consist of two separate GANs and two data domains. Each GAN transforms data from one domain to the other. In addition, there is a cycle consistency loss, which enforces that applying the generators of both GANs sequentially maps back to the original input. CycleGAN and its derivatives have been used for style transformations \cite{brunner2018symbolic,karras2019style,tmenova2019cyclegan}. Since segmentation is a type of style transformation \cite{thambawita2022singan}, CycleGANs can also be used for segmentation in both a supervised \cite{huo2018adversarial} and unsupervised environment \cite{chen2018semantic,ihle2019unsupervised}. The downside of these approaches is that they are not end-to-end, as they only segment the data but do not extract features directly.

Many object-centric representation learning approaches demonstrate their strength on benchmarks such as Pascal VOC \cite{everingham2010pascal} or MoVi \cite{ghorbani2020movi}. Here, objects are large, visually distinctive, and embedded in complex, textured backgrounds. These settings allow for extra cues such as temporal continuity in video, strong background-foreground contrast, or semantic diversity. This helps object-centric representation learning methods to keep objects apart. In contrast, the datasets that motivate our work contain up to hundreds of small, nearly identical objects in low-contrast environments. Instance segmentation becomes challenging in such real-world conditions. In addition, real-world datasets contain richly textured backgrounds \cite{seitzer2022bridging}. 

For simplicity, this work attempts to instance segment real-world datasets with homogeneous backgrounds. Such datasets can, for example, be found particularly in microscopy images. Here, small objects are often overlooked, and visually similar but distinct objects may be incorrectly grouped. We will show that our approach can learn robust object-centric representations in such scenarios, while existing architectures struggle. Furthermore, we show that CycleGANs correctly identify object counts even when test-time properties, such as object spacing and object count, differ from the training data. 

The main contributions of this manuscript are:
\vspace{-0.5\baselineskip}
\begin{itemize} 
    \item We propose a novel, fully differentiable CycleGAN framework \cycleGANli, which mutually transforms an image and a list domain for unsupervised object-centric representation learning.
    \item We match state-of-the-art performance on synthetic datasets, while at the same time being the only approach that can reliably solve a more complex real-world dataset consisting of large sets of visually similar objects.
    \item We show that our method maintains its performance across different image dimensions, allowing a single trained model to scale without retraining.
\end{itemize}

%% file: sec/2_related_work.tex
\section{Related Work}
\label{sec:related_work}

\subsection{Generative Adversarial Networks}

    Generative Adversarial Networks (GANs) can be combined with lists (\ie~tables) \cite{xu2018synthesizing,park2018data}. These works have treated the entire dataset as a single large list. In contrast, our approach treats every object as a single entry in the list, making the entire list analogous to a single image rather than the full dataset.
    
    CycleGANs were originally introduced for translating between two image domains~\cite{zhu2017unpaired}. Since then, CycleGAN architectures have been applied to other data modalities, such as sound \cite{kaneko2019cyclegan} and one-dimensional signals \cite{easthope2024paired}. Additionally, CycleGANs can facilitate cross-domain transfer. For instance, Gorti and Ma proposed a CycleGAN architecture capable of cross-modal transformations between text and images \cite{gorti2018text}.

\subsection{Object-Centric Representation Learning}

    Previous work on object-centric representation learning using GANs focuses on image merging and segmentation. Various techniques have been proposed to extract object masks and separate them from the background \cite{arandjelovic2019object,abdal2021labels4free} or to undo occlusion \cite{kortylewski2021compositional}. Another approach is object-wise image generation \cite{van2020investigating}, where individual images are generated for each object and then aggregated. Models that explicitly encode transformation parameters for scene modification have also been proposed \cite{nguyen2020blockgan,ehrhardt2020relate}. Except for \cite{ehrhardt2020relate}, these models assume a fixed number of objects. 

    Spatial-attention models have been introduced by Eslami \textit{et al.} through the Attend, Infer, Repeat (AIR) framework \cite{eslami2016attend}. AIR recovers the objects iteratively using a recurrent neural network, which is intractable. Several derivatives exist that mitigate this problem by following a mean-field approach, such as GMAIR \cite{zhu2022gmair} and SPAIR \cite{crawford2019spatially}, where an image is divided into subregions as proposed by YOLO and others \cite{redmon2016you,liu2016ssd,redmon2018yolov3}. While these approaches are fast and scalable, their mean-field nature means that the recovered objects cannot be conditioned on each other. In addition, they struggle with objects that are larger than the subregion.

    Slot attention-based models infer object slots through attention mechanisms \cite{locatello2020object}. SPACE \cite{lin2020space} combines a spatial attention model for the foreground with a scene-mixture model for the background. SLATE \cite{singh2021illiterate} couples a slot-based encoder with a transformer decoder, while LSD \cite{jiang2023lsd} uses an unsupervised compositional conditional diffusion model that employs a latent diffusion framework conditioned on object slots. GOLD \cite{chen2025gold} improves object-centric learning by disentangling intrinsic and extrinsic attributes. Scene-mixture models assign pixels to slots, ideally representing objects or background, but face the ``slot-decoding dilemma'' \cite{singh2021illiterate}, requiring iterative slot refinement \cite{engelcke2021genesis,locatello2020object}, which is computationally expensive. SuPAIR \cite{stelzner2019faster} addresses this with sum-product networks, while SPOT \cite{kakogeorgiou2024spot} integrates self-training and sequence permutation strategies. To simplify slot extraction, SAMP \cite{patil2024samp} employs convolutional and max pooling layers. Despite these advancements, models still struggle with highly textured objects \cite{lowe2022complex}, as well as small or numerous objects \cite{kakogeorgiou2024spot}.

    Another line of work has explored object discovery through information-theoretic principles. For instance, Wolf \textit{et al.} proposed an inpainting-based segmentation model that minimizes information gain between partitions to segment densely packed cells in microscopy images \cite{wolf2020inpainting}. More recently, complex-valued neural networks \cite{lowe2022complex}, inspired by the temporal correlation hypothesis from neuroscience, have emerged as a novel paradigm for object discovery. These models leverage complex-valued representations to capture temporal dependencies, offering an alternative to conventional segmentation approaches.

%% file: sec/3_methodology.tex
\section{Methodology}

Our model architecture is based on the CycleGAN \cite{zhu2017unpaired}. CycleGANs learn a mapping between two different domains given unpaired training samples. In contrast to the original CycleGAN, which uses solely images, we work with two different data modalities. The goal is to convert an image $I \in \mathbb{R}^{h\times w\times c}$ to a list $L \in \mathbb{R}^{k\times (3+m)}$, where $m = |\eta|$ denotes the dimensionality of the feature vector, and back by enforcing cycle-consistency. A list entry $l\in \mathbb{R}^{(3+m)}$ contains the $x$ and $y$ coordinates of an object, a scalar value $\alpha$ that indicates the probability that an object exists, and a feature vector $\eta$ encoding object properties.

In addition, a generated list or image is shown to a modality-specific discriminator $D_M, M \in \lbrace I,L \rbrace$, where $M$ is the image or list modality. The discriminator learns to distinguish between generated and real data. This adds a minimax game to the overall optimization problem, where the generator learns to fool the discriminator. The complete procedure is illustrated in \fig{fig:0_train}{}. Network implementation details and hyperparameter values are provided in the Supplementary Information.

\subsection{Architecture}

\begin{figure*}[t!]
    \centering
    \includegraphics[width=1.0\linewidth]{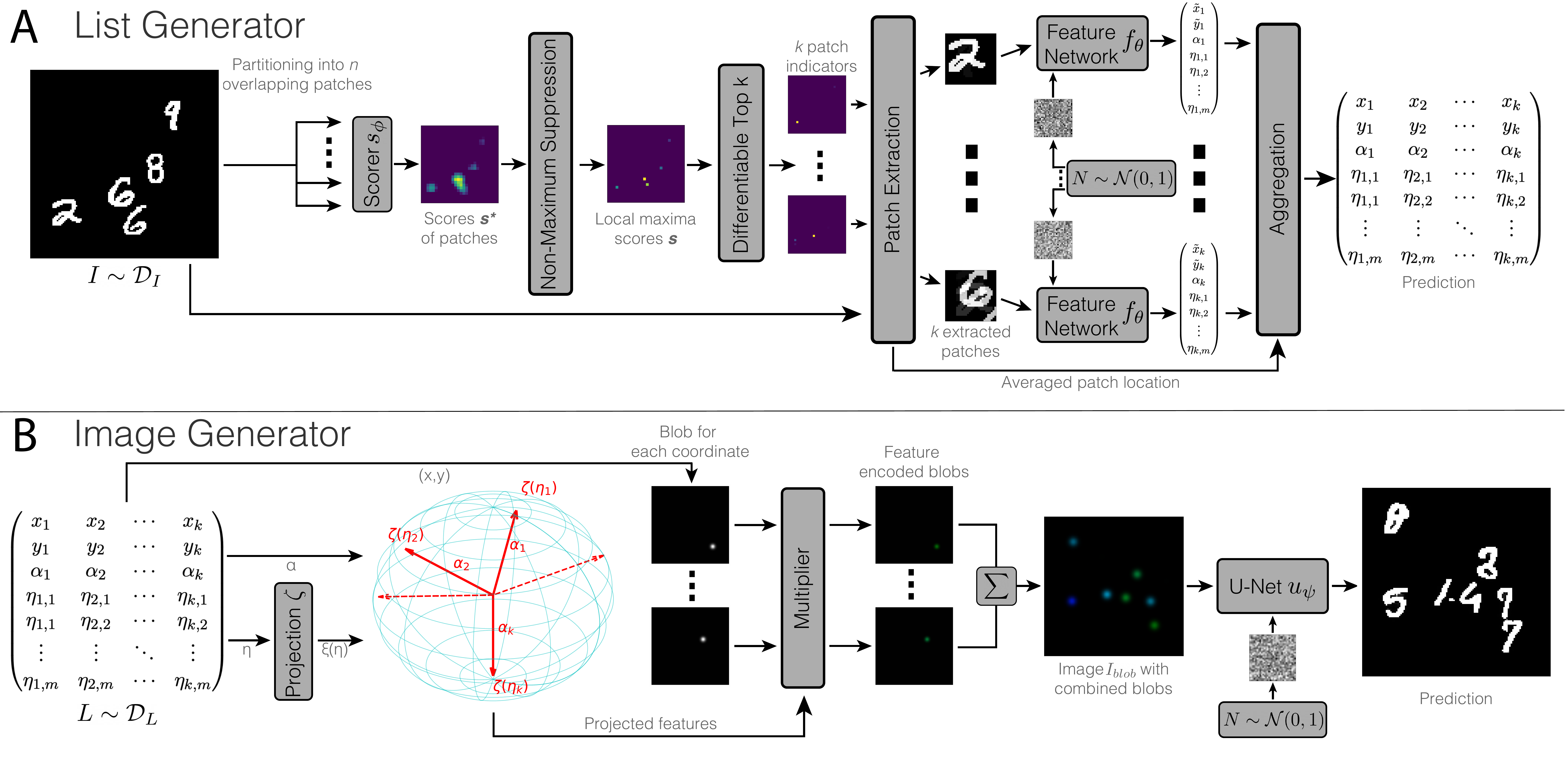}
    \caption{Network architectures used in this work.
    \textbf{(A)} List generator architecture, which transforms images into lists. First, the image is divided into patches and a score function is used, which predicts for each patch how likely it is to contain an object. After non-maximum suppression, a differentiable top-k operator detects k patches that contain objects. The feature network then extracts the object's features.
    \textbf{(B)} Image generator architecture, which transforms lists into images. First, the list is transformed into an image based on its location and features. For that, each feature vector $\eta_i$ is projected ($\zeta$) onto a unit hypersphere. Afterwards, the vector is scaled by the object existence probability $\alpha$. Then, a similar approach as presented in BlobGAN is used \cite{epstein2022blobgan} to convert the projections into an image. After adding a noise channel to allow for richer backgrounds, a U-Net transforms the style of the image.}
    \label{fig:1_architecture}
\end{figure*}

Since there are two different data modalities present, two separate generator architectures are required. Both generators are illustrated in \fig{fig:1_architecture}{A} and \fig{fig:1_architecture}{B}. 

\subsubsection{List Generator}

The list generator $G_L(\cdot;\theta,\phi)$ maps $I$ to $L$. This network consists of two key steps. In the first step, patches of interest are extracted. In the second step, the features of each object are predicted.

The list generator divides each image into $n$ overlapping patches $p_i \in \mathbb{R}^{p_h\times p_w\times c}$ with a fixed stride. The patch size is chosen such that the biggest object of interest in the scene fits within. Overlapping the patches is crucial for a more precise location estimate. A convolutional score network $s_\phi$ then processes all patches in parallel. For each $p_i$, a score is returned that indicates the importance of the patch to be used for feature extraction. Due to the convolutional nature of $s_\phi$, neighboring patches get a similar score. To mitigate the issue of marking patches for the same object multiple times with a high score, non-maximum suppression (NMS) is applied. The $k$ patches of the highest interest are then extracted using an adapted differentiable top-k algorithm \cite{cordonnier2021differentiable}. Each of the top $k$ patches is then independently processed by a convolutional feature network $f_{\theta}$. A noise channel is incorporated in $f_{\theta}$. This ensures $D_L$ cannot exploit the fact that identical patches yield the same feature description. The feature network outputs one entry in the list for each patch. As the global position of the patch is unknown to the network, it can only determine the offset of the object relative to the patch's center. The extracted offset is clipped to the stride of the initial partition. Then, the final location prediction is achieved by adding the patch location and the extracted offset.

Unlike other approaches considered in this work, our list generator is fully convolutional with respect to the input image and is therefore not tied to a fixed image size. This allows us to train on small training images and generate lists for larger test images. The number of extracted patches $k$ is a hyperparameter that can be tuned according to the maximum expected number of objects within an image. 

\subsubsection{Image Generator}

The architecture of the image generator $G_I$ is a modification of~\cite{epstein2022blobgan}. Unlike their approach, the mapping from our list $L$ to an initial image $I_{blob} \in \mathbb{R}^{h\times w\times (m+1)}$ involves no learnable parameters. For each list entry $l_i$, an isotropic Gaussian blob of fixed variance is generated at the coordinates $(x_i,y_i)$. Each feature vector $\eta_i$ is mapped onto a unit hypersphere using a fixed projection function:
\begin{equation}
    \zeta(\eta_i) = 
    \left[
        \cos(\pi \eta_{i,1}),
        \ldots,
        \prod_{j=1}^{m-1}\sin(\pi \eta_{i,j})\cos(\pi \eta_{i,m}),
        \prod_{j=1}^{m}\sin(\pi \eta_{i,j})
    \right],
\end{equation}
which is then scaled by $\alpha_i$. This scaled projection defines the channel-wise amplitudes of the corresponding blob:
\begin{equation}
I_{\text{blob}}(x, y, d) = \sum_{i=1}^k \alpha_i \, \zeta\left(\eta_{i}\right)_{d} \, e^{\left( -\frac{(x - x_i)^2 + (y - y_i)^2}{2\sigma^2} \right)},
\end{equation}
where $x$ and $y$ are the spatial coordinates and $d$ is the channel of the resulting tensor. $I_{blob}$ is then processed by a U-Net \cite{ronneberger2015u} $u_\psi$. An image has many more degrees of freedom, and hence its complexity is much greater than that of a list. To ensure that the image discriminator $D_I$ cannot easily exploit this fact, $u_\psi$ additionally receives a noise tensor. 

\subsubsection{Discriminators}

To distinguish between synthetic and real images, a PatchGAN is used for the image discriminator $D_I$ \cite{zhu2017unpaired,isola2017image}. For the list discriminator $D_L$, two conditions are required: (1)~Relations between elements have to be captured. (2)~The ordering of list elements must have no effect. The first condition provides feedback on inter-object relations. The latter is important, as a list describes the same scene irrespective of the order of its elements. The list discriminator satisfies both conditions by combining attention modules \cite{vaswani2017attention} with a final global average pooling layer. Each list element is treated as a token without positional encoding. The complete list discriminator architecture is a simplified version of the Point Transformer architecture by Zhao \textit{et al.} \cite{zhao2021point} and described in the Supplementary Information.

\subsection{Training}

Both generators learn to map to an output that resembles a sample drawn from the target distribution $\mathcal{D}$ by minimizing an adversarial loss $\mathcal{L}_{Dis}$ \cite{goodfellow2014generative}. The discriminator $D_M$ tries to distinguish between synthetic and real samples and aims to maximize $\mathcal{L}_{Dis}$. For both modalities, the least squares approach as presented by Mao \textit{et al.} was used \cite{mao2017least}. To enforce an invertible mapping, a cycle-consistency loss is introduced. The quality of a cycled image is assessed by calculating the mean absolute error (MAE) with respect to the input image. For the list cycle loss, an element-wise loss is defined. As the lists are unordered, a bijective mapping for each object between two lists is required. The mapping is created by formulating it as a linear sum assignment problem. In this work, the assignment problem was solved by a modified version of the Jonker-Volgenant algorithm \cite{jonker1988shortest,crouse2016implementing}. The cost associated with a pair of elements depends on both the location and the probability of its presence:
\begin{equation}
\begin{split}
    \mathcal{C}_{Obj}(l_i,l_j) = \frac{\alpha_i}{w^2} \left\| \begin{pmatrix} x_i \\ y_i \end{pmatrix} - \begin{pmatrix} x_j \\ y_j \end{pmatrix} \right\|^2 
     + \lambda_{pres} \big( \alpha_i - \alpha_j \big)^2.
\end{split}
\end{equation}

The image width $w$ normalizes the spatial distance while the scaling factor $\lambda_{pres}$ ensures that for the comparison, object presence is of sufficient importance. After object assignment, the cycled list $l^{Cyc}$ is reordered such that entry $i$ matches entry $i$ of the input list. The matched entries are compared using the following loss function:

\begin{equation}
\begin{split}
    \mathcal{L}_{Cyc_L} = \mathbf{E}_{L} \Bigg\{ 
    \frac{1}{k} \sum_{i=1}^k \Bigg[\mathcal{C}_{Obj}(l_i,l_i^{Cyc}) +
         \frac{\big\| \eta_i - \eta_i^{cyc}  \big\|^2}{m} \Bigg] 
    + \lambda_{loc} \frac{\left\| s_\phi(p) - t_p \right\|^2  }{|t_p|}\Bigg\}, 
\end{split}
\end{equation}

with $t_p$ being the target for the score network calculated from the locations of the objects, and $s_\phi(p)$ is the output of the score network for all patches. The last term serves as an additional gradient for $s_\phi$. During training, it is crucial that $s_\phi$ extracts the right patches early in training. Its importance is tuned by the scalar hyperparameter $\lambda_{loc}$. 

Due to the two modalities, different weightings for losses of the list, image cycle, and adversarial losses are required. The loss function of the generator can be summarized as 
\begin{equation}
\begin{split}
    \mathcal{L}_{Gen} = 
    \lambda_{Dis_L} \mathcal{L}_{Dis_L} + 
    \lambda_{Dis_I} \mathcal{L}_{Dis_I}
    + \lambda_{Cyc_I} \mathcal{L}_{Cyc_I} 
    + \lambda_{Cyc_L} \mathcal{L}_{Cyc_L},
\end{split}
\end{equation}

where the first terms correspond to the adversarial losses of the two domains, and each $\lambda$ is a scalar value that weights the respective loss component. The hyperparameters for patch size $p_h,p_w$ and the stride for patch extraction are adjusted according to the dataset used. In this work, the patch size is restricted to the case where $p_h=p_w$ and the stride is set to $\left \lceil{\frac{p_w}{8}}\right \rceil $ for simplicity. The parameter $\sigma$, which determines the size of generated blobs in $G_I$ is fixed to $\frac{p_w}{10}$. To more effectively deal with objects on the boundary of the image, we optionally pad the input with $n_{pad}$ pixels. In addition, the dimensionality of $\eta$ depends on the specific dataset. All other hyperparameters are equivalent for all experiments. To enhance training stability, both discriminators employ a linear warm-up schedule for their learning rates and apply spectral normalization \cite{miyato2018spectral}.

\subsection{Datasets}

We benchmark on four datasets (more details on datasets in Supplementary Information): 

\vspace{-1em}
\paragraph{\tetris \cite{multiobjectdatasets19}:} $35\times35$ images with three non-overlapping tetrominoes of random colors and positions. The list has three entries.

\vspace{-1em}
\paragraph{\sprites:} Up to ten colored sprites (squares–stars) on a $128\times128$ blue background, positions drawn with Poisson-disk sampling. The list has ten entries.

\vspace{-1em}
\paragraph{\mnist \cite{deng2012mnist}:} Up to ten handwritten digits per $128\times128$ image, possibly overlapping, locations are Poisson-disk sampled. The list has ten entries.

\vspace{-1em}
\paragraph{\cells \cite{Azmi2023blood}:} A real-world dataset of $128\times128$ microscope patches of blood containing many, visually similar, low-contrast cells. The list has ten entries.

For all datasets, the list domain is sampled independently of the images and defines a prior over object presence, location, and features. Feature vectors are sampled uniformly at random. For synthetic datasets, presence and locations follow the same generative priors as for the images. For \cells, $k=10$ is set as an upper bound for the number of cells in a $128\times128$ patch. Presence is sampled from $\mathrm{Bernoulli}(0.5)$, and locations are sampled by Poisson-disk sampling with a minimum center distance of $25$ pixels.

%% file: sec/4_experiments.tex
\section{Experiments}
\label{sec:experiments}

We compare \cycleGANli with four other recent approaches, where each covers a different technique used in unsupervised object-centric representation learning. SPACE \cite{lin2020space} is the most closely related to \cycleGANli as it is also spatial-attention based. At the same time, it incorporates a scene-mixture approach. Further, two slot-based models, LSD \cite{jiang2023lsd} and SLATE \cite{singh2021illiterate} are presented. Finally, SPOT \cite{kakogeorgiou2024spot}, a self-supervised slot-based model with self-training and sequence permutations, has been trained for comparison. An ablation study of \cycleGANli can be found in Supplementary Information online, where the effects of the discriminators, cyclic losses, differentiable top-k and warmup are investigated. Furthermore, the Supplementary Information shows the effect of non-homogeneous backgrounds and the patch size on the performance of \cycleGANli.

\subsection{Qualitative Comparison}

\begin{figure*}[t]
  \centering
  \includegraphics[width=1.0\linewidth]{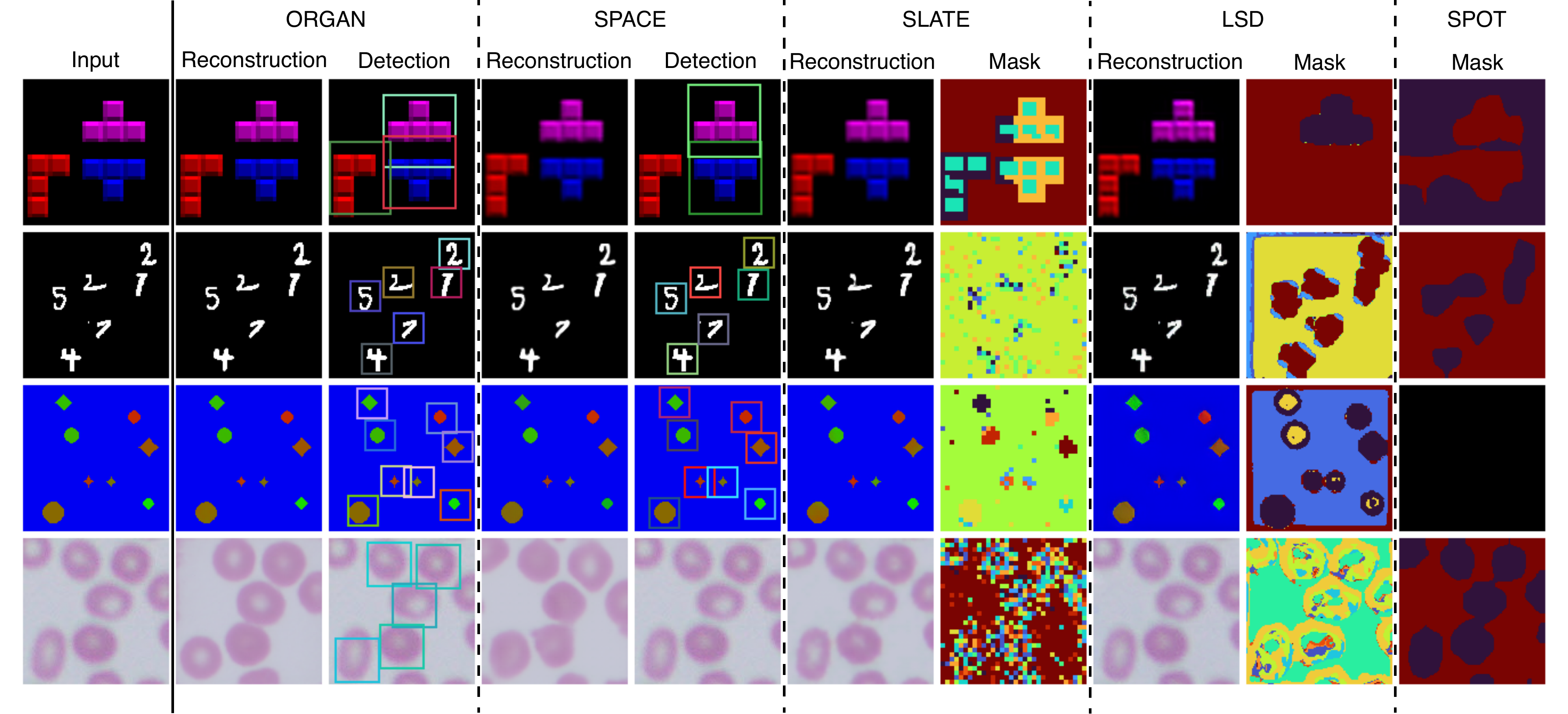}
  \vspace{-1em}
   \caption{Examples of object detection performance for four different datasets. We show performance for ORGAN (ours), SPACE \cite{lin2020space}, SLATE \cite{singh2021illiterate}, LSD \cite{jiang2023lsd} and SPOT \cite{kakogeorgiou2024spot}. We present both the reconstruction and the object detection in the form of fixed bounding boxes for ORGAN and SPACE. The color code of the bounding box represents the extracted feature vector. For SLATE, LSD, and SPOT we present the recovered slot masks (color-coded). SPOT does not allow for image reconstruction. }
   \label{fig:comparison}
\end{figure*}

To qualitatively assess the performance of all models, their reconstructions were compared. In addition, the masks for the different slots or the bounding boxes around the center of the predicted objects were visualized. The results can be seen in \fig{fig:comparison}{}.

\cycleGANli gives a detailed reconstruction of the input images. In addition, it manages to precisely detect most objects across all four datasets. 

SPACE shows accurate reconstructions for the synthetic datasets. It reliably detects separate objects in \sprites and \mnist. For \tetris, it assigns certain tetrominoes (in our case, red ones) to the background. For \cells, SPACE fails to detect any cell as an object, resulting in reconstructions generated solely by its background module. The low contrast of the dataset makes it difficult to separate foreground from background. 

SLATE achieves precise reconstructions across all datasets but inferior mask assignments, as separate objects are assigned to the same slots. For example, in \tetris, it separates the block boundaries from its centers. The model even allocates multiple slots to a single object without unique associations, as is done in \sprites. In addition, for \cells and \mnist, the segmentation yields only limited structural detail.

LSD separates the foreground from the background across all synthetic datasets. However, in most cases, it assigns all distinct objects to a single slot while reserving another slot for the background. In the \sprites dataset, LSD partially succeeds in separating green objects while merging those of other colors into the same slot. In contrast, in the \cells dataset, LSD struggles to distinguish the foreground from the background, likely because the cell centers share the same color as the background, leading to their misallocation as part of it.

SPOT demonstrates the ability to separate foreground from background in most datasets; however, its masks appear coarser than those produced by other models. The model is most challenged by the \sprites dataset, which consists of multiple small objects. Here, SPOT over-segmented the region, failing to isolate individual objects. Instead of assigning distinct slots to separate objects, SPOT tends to merge multiple entities into a single slot. \ifAIAI\else As shown in the Supplementary Information, t\fi his behavior is likely due to the encoder’s slot-attention mechanism, which struggles to generate sufficiently distinct slot representations.

\subsection{Quantitative Comparison}

\begin{table*}[t]
\centering
\scriptsize
\caption{F1-scores (in \%) and training/inference times (hours/ms) for all models on 128$\times$128 images.}
\label{tab:comparison}
\resizebox{\linewidth}{!}{
\begin{tabular}{lcc|cc|cc|cc}
\toprule
 & \multicolumn{2}{c|}{\tetris} 
 & \multicolumn{2}{c|}{\mnist} 
 & \multicolumn{2}{c|}{\sprites} 
 & \multicolumn{2}{c}{\cells} \\
Method 
 & F1 & Train/Infer
 & F1 & Train/Infer 
 & F1 & Train/Infer 
 & F1 & Train/Infer \\
\midrule
ORGAN 
 & \textbf{89.2$\pm$20.5} & 44.0 / \textbf{2.5}
 & 86.1$\pm$3.9 & 39.0 / 3.5
 & 88.1$\pm$2.3 & 36.0 / 4.0
 & \textbf{77.9$\pm$8.6} & 31.5 / \textbf{3.4} \\

SPACE 
 & 76.0$\pm$15.7 & 13.5 / 17.8
 & \textbf{88.7$\pm$4.7} & 16.5 / 15.0
 & \textbf{88.5$\pm$4.3} & 4.08 / 29.6
 & 0.0$\pm$0.0 & 13.0 / 15.0 \\

SLATE 
 & 30.3$\pm$24.8 & \textbf{6.75} / 3.4
 & 5.8$\pm$7.6 & \textbf{0.75} / \textbf{3.4}
 & 12.4$\pm$10.6 & \textbf{0.75} / \textbf{3.0}
 & 46.6$\pm$9.3 & \textbf{0.75} / 3.5 \\

LSD 
 & 20.0$\pm$39.8 & 7.5 / 8.0
 & 5.7$\pm$28.0 & 16.0 / 4.0
 & 6.4$\pm$17.7 & 11.5 / 7.0
 & 53.9$\pm$11.9 & 10.5 / 7.0 \\

SPOT 
 & 13.8$\pm$44.6 & 10.0 / 37.5
 & 5.2$\pm$43.0 & 10.0 / 36.3
 & 0.5$\pm$21.2 & 10.0 / 32.8
 & 24.8$\pm$0.0 & 10.0 / 32.8 \\
\bottomrule
\end{tabular}}
\end{table*}

The models were quantitatively evaluated using the F1-score for object detection accuracy, along with measurements of training and inference time (Table~\ref{tab:comparison}). For SLATE, LSD, and SPOT, object locations were extracted as the center of mass of each slot’s segmentation mask. For the \cells dataset, 50 images were annotated by an expert to provide ground-truth object positions. All training and inference experiments were conducted on an NVIDIA RTX 2080 Ti GPU.
Across all presented datasets, the evaluated slot-based models (SLATE, LSD, SPOT) do not reliably detect distinct objects. In contrast, SPACE and \cycleGANli achieve substantially higher detection performance. While SPACE trains more than twice as fast as \cycleGANli, inference with \cycleGANli is approximately five times faster. Since only SPACE and \cycleGANli consistently detect distinct objects, subsequent experiments are restricted to these two models.

\subsection{Expressiveness of Feature Space}

\begin{figure}[t]
    \centering
    \includegraphics[width=1.0\linewidth]{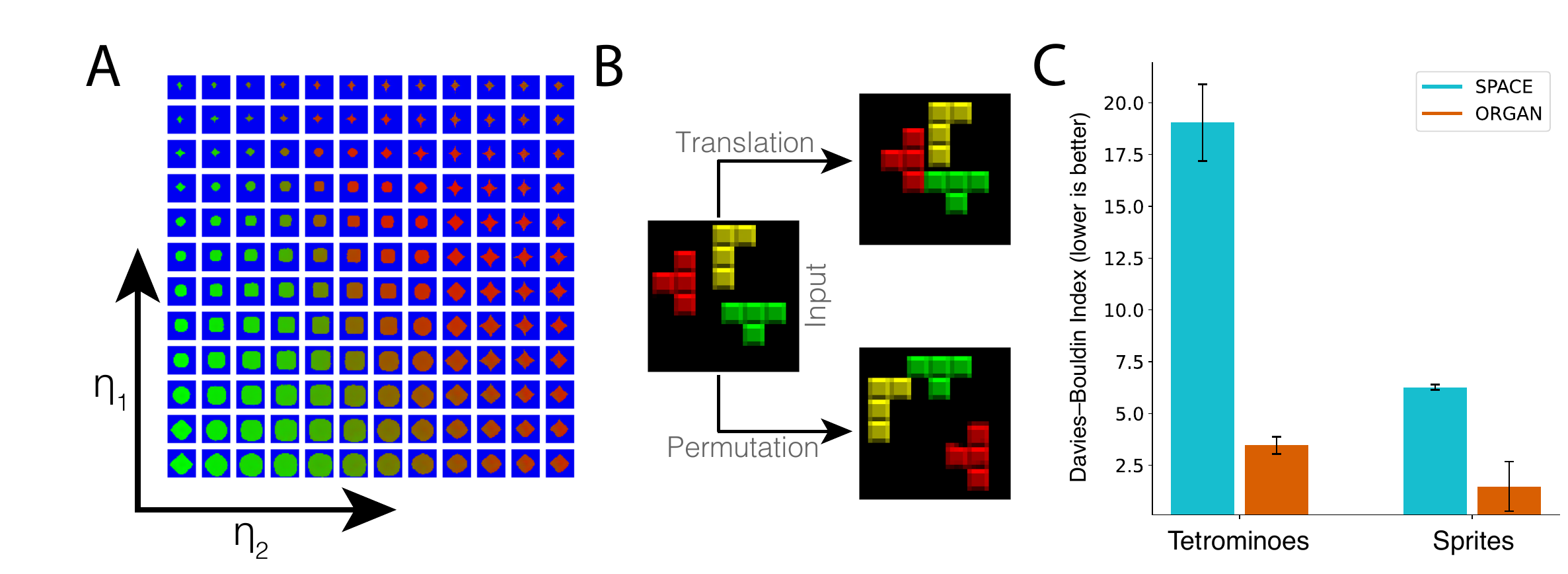}
    \vspace{-1em}
    \caption{Latent space quality of \cycleGANli. 
    \textbf{(A)} The latent space as encoded by $\eta$ is presented for the \sprites dataset. Two of the three dimensions are sampled at 12 equally spaced values across the feature space. The third dimension is kept fixed at its center. Each combination of feature values is given to the image generator. The corresponding objects are generated and visualized in separate sub-images. The feature space encodes the color, shape, and size of the object.
    \textbf{(B)} An input image of the \tetris dataset is cycled, while the list elements are modified. Top: All objects were moved closer to the center. Bottom: The object properties were swapped between the three objects counterclockwise.
    \textbf{(C)} To compare how well object clustering works on the feature space, the Davies–Bouldin Index \cite{davies1979cluster} was calculated for \cycleGANli and SPACE, which detected distinct objects for \sprites and \tetris. Our method achieves superior separability.}
    \label{fig:latent}
\end{figure}

The \sprites dataset was generated using a three-dimensional feature vector where each dimension corresponds to either color, scale, or shape. Therefore, it is possible to contrast the true feature space to the feature space of \cycleGANli. A two-dimensional slice through the feature space is shown in \fig{fig:latent}{A}. The feature space shows changes in object size, shape, and color. This indicates that \cycleGANli is capable of disentangling the location and object properties. This capability is further demonstrated in \fig{fig:latent}{B}, where an input image of \tetris is first transformed into a list, then the list is modified, and finally cycled back into an image.

We were further interested in how consistently similar objects are placed within the extracted feature space. For this, we used a modified \sprites dataset with three shapes, colors, and sizes (27 object types), as well as the \tetris dataset. Feature vectors were grouped by object type, and group separability was assessed using the Davies–Bouldin Index \cite{davies1979cluster}. Results are shown in \fig{fig:latent}{C}, where \cycleGANli significantly outperforms SPACE, indicating a more structured and separable feature space.

\subsection{Detection Accuracy in Large Scenes}

\begin{figure}[t]
    \centering
    \includegraphics[width=1.0\linewidth]{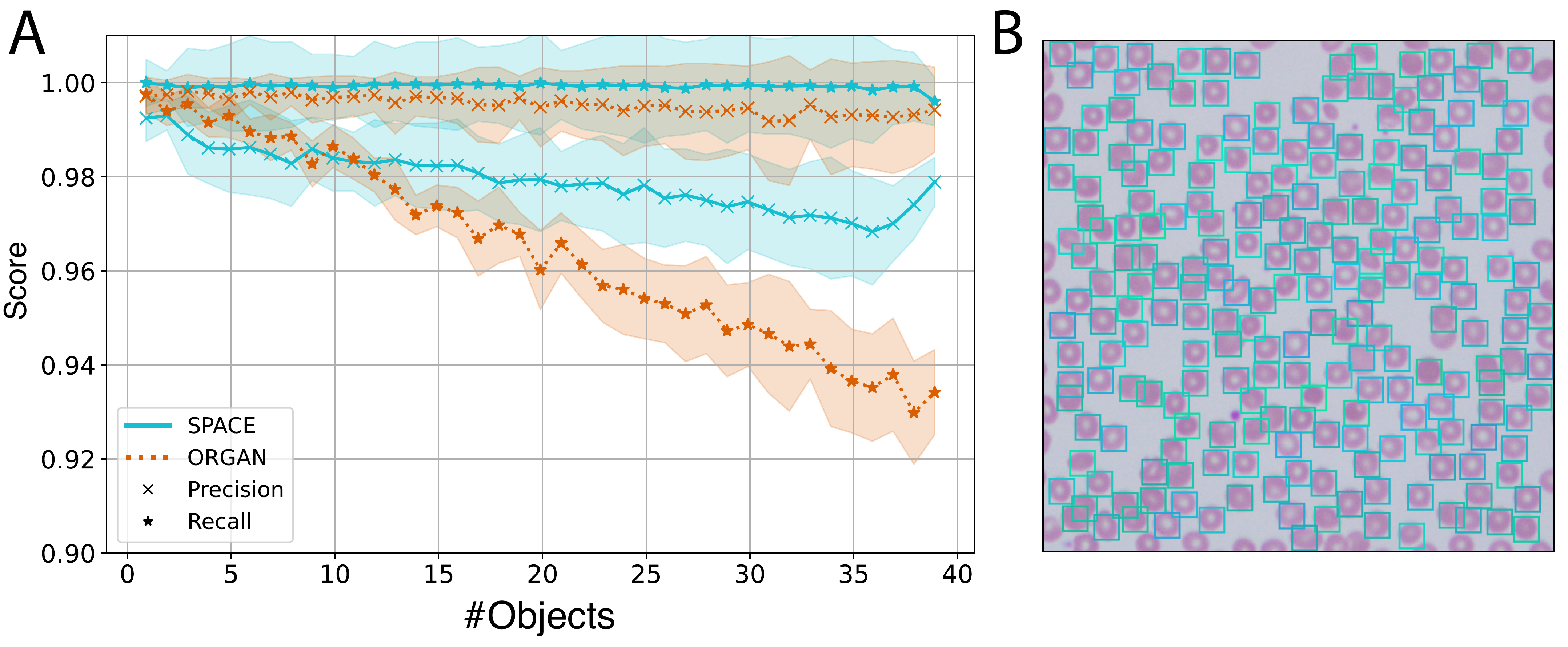}
     \caption{Analysis of the generalization capabilities of the object detection accuracy. 
     \textbf{(A)} Recall and precision are plotted for both \cycleGANli and SPACE for \sprites. The architecture of SPACE can only be applied to a fixed input size (here $256 \times 256$). \cycleGANli was trained on $128 \times 128$ and applied to $256 \times 256$. This training approach gives SPACE an advantage because it is trained directly at the evaluation resolution. Yet, \cycleGANli outperforms in precision, while SPACE has a higher recall.
     \textbf{(B)} \cycleGANli can also generalize to much larger images. Trained on patches of size $128 \times 128$, prediction results on a $768 \times 768$ image of the \cells dataset are presented. The color code of the bounding box illustrates differences in extracted features $\eta$.}
     \label{fig:large_scale}
\end{figure}

In order to investigate how well \cycleGANli works on images of different sizes and with many more objects than present during training, we trained it on the standard \sprites dataset with an image size of $128 \times 128$ pixels. Here, only models that exhibited sustained improvements in cycle loss by the 100$^\text{th}$ epoch were retained (acceptance rate: 75~\%). \cycleGANli was evaluated on a new dataset of size $256 \times 256$ with up to 39 objects. Network performance was measured using precision and recall metrics. The results are shown in \fig{fig:large_scale}{A}. SPACE was used to compare the performance of our architecture. As SPACE does not allow for images of different sizes for training and prediction, it was trained directly on $256 \times 256$ inputs.

When comparing precision and recall for SPACE and \cycleGANli (see \fig{fig:large_scale}{A}), no distinction between the two models can be made when only a few objects are present. However, as the number of objects in an image increases, the recall metric drops for \cycleGANli (recall at 39 objects for \cycleGANli: 93.4\%, SPACE: 99.6\%). A likely factor behind the recall drop is the NMS, which can cause nearby objects to suppress each other, causing missed detections. Still, \cycleGANli outperformed SPACE across the board when considering the precision metric (at 39 objects for \cycleGANli: 99.4\%, SPACE: 97.9\%).

To further evaluate \cycleGANli on real-world applications, it was trained for \cells with images containing $128 \times 128$ pixels and tested on a large image of $768 \times 768$ pixels. The result is shown in \fig{fig:large_scale}{B}. Three experts provided the ground-truth annotations, with an inter-annotator F1 of 97.7\%; the model scored an average F1 of 75.0\% compared to the experts. This test illustrates the strong generalization capabilities of \cycleGANli. The small patches seen during training contain partially cut-off cells on the boundary. Nevertheless, the approach does not seem affected by this shortcoming of the training data and successfully captures a broader scene.

When models are restricted to a fixed input size, input images are usually sliced into multiple, potentially overlapping parts. As our model allows us to scale to different input sizes regardless of the size of the image during training, the need to slice large inputs is unnecessary. This is beneficial, as slicing has an inherent problem of boundary effects.

%% file: sec/5_conclusion.tex
\section{Conclusion}
\label{sec:conclusions}

Our approach was the only one tested that detected separate entities on a low-contrast image of cells. We showed that the learned feature mapping allows for an easier distinction of different object types compared with non-GAN approaches like SPACE. We attribute these performance gains to two key factors: First, GANs produce sharper reconstructions compared to VAEs \cite{wang2021generative}; second, the cyclic architecture of \cycleGANli provides additional training feedback for the generators, which can be effectively leveraged. Hence, our approach effectively addresses a common criticism of object-centric representation learning: the purported inability of GAN-based methods to handle large numbers of objects \cite{yang2022promising,yang2024benchmarking}.

Debate persists over whether object slots should be conditioned on previously recovered objects. Crawford and Pineau add such a condition to prevent duplicate detections, but it is intractable and scales poorly \cite{crawford2019spatially}. SPACE replaces it with a mean-field, parallel detector that is more scalable yet ignores inter-object dependencies \cite{lin2020space}. \cycleGANli sidesteps both issues: a differentiable top-k locator yields tractable proposals, while the GAN discriminator with self-attention and non-maximum suppression penalizes duplicates, letting slots remain mutually dependent \cite{vaswani2017attention}. This combination keeps the model scalable without the weight-growth or sequential bottlenecks that hamper current approaches.

Despite its effectiveness, \cycleGANli has several limitations. \cycleGANli does not aim to reconstruct complex non-homogeneous backgrounds and may therefore struggle with datasets that have such backgrounds, as there currently is no information pathway in the model to encode background properties. Due to this reason, datasets such as COCO \cite{lin2014microsoft} are not suitable for \cycleGANli.
In addition, due to its GAN-based architecture, \cycleGANli can be inherently unstable during training and care needs to be taken during the minimax optimization of the GANs. 
Furthermore, the patch-based approach means that processing any object larger than a patch, or two objects that lie too close together across a patch boundary, may be missed.
Finally, although \cycleGANli does not require object-level annotations, it is not fully assumption-free by relying on user-specified list distributions. While these distributions need not be specified precisely, they must remain sufficiently consistent with the underlying data distribution to ensure stable training and reliable performance. Tuning the corresponding hyperparameters, such as the size of the list, is not unique to \cycleGANli and is also required for other state-of-the-art approaches.

%% file: sec/6_acknowledgements.tex
\ifAIAI
\else
\subsection*{Acknowledgements}
\fi

This research was supported by ETH Zürich, the Swiss National Science Foundation (SNSF) [project number 10.001.282], and the Swiss Data Science Center (SDSC). The authors also thank Peter Littlewood and J\'anos V\"or\"os for their continued support. S. J. Ihle is supported by the AI in Science Fellowship, a program of Schmidt Sciences.

\ifAIAI
\else
\subsection*{Availability Statement}
\fi
\label{sec:availability}

\ifAIAI\else The code is \fi publicly available on GitHub: \url{https://github.com/Hullimulli/ORGAN}. 